\documentclass[10pt,journal,compsoc]{IEEEtran}
\usepackage{footnote}
\makesavenoteenv{table}
\usepackage{tablefootnote}

\usepackage{multirow}
\usepackage{tabularx}
\usepackage{siunitx} 
\usepackage{graphicx}
\usepackage{ragged2e}
\usepackage[colorlinks,linkcolor=red]{hyperref}
\usepackage{xcolor}

\definecolor{amber}{rgb}{1.0, 0.49, 0.0}

\ifCLASSOPTIONcompsoc
  \usepackage[nocompress]{cite}
\else
  \usepackage{cite}
\fi
\ifCLASSINFOpdf
\else
\fi
\begin{document}
%
\title{OmniCity: Omnipotent City Understanding with Multi-level and Multi-view Images}
%
%
%
%

\author{Weijia~Li,
        Yawen~Lai,
        Linning~Xu,
        Yuanbo~Xiangli,
        Jinhua~Yu,
        Conghui~He,
        Gui-Song~Xia
        and~Dahua~Lin
\IEEEcompsocitemizethanks{
\IEEEcompsocthanksitem W. Li and J. Yu are with Sun Yat-sen University, China. \\
E-mail: liweij29@mail.sysu.edu.cn, leonardyu1126@gmail.com.
\IEEEcompsocthanksitem Y. Lai is with Peking University, China. \\
Email: alanlyawen@pku.edu.cn.
\IEEEcompsocthanksitem L. Xu, Y. Xiangli and D. Lin are with The Chinese University of Hong Kong, Hong Kong, China. \\
E-mail: \{xl020,xy019,dhlin\}@ie.cuhk.edu.hk.
\IEEEcompsocthanksitem C. He is with SenseTime Research, China. \\
E-mail: heconghui@sensetime.com.
\IEEEcompsocthanksitem G.-S. Xia is with Wuhan University, China. \\
E-mail: guisong.xia@whu.edu.cn.
}
}

\IEEEtitleabstractindextext{%
\begin{abstract}
\justifying
This paper presents OmniCity, a new dataset for omnipotent city understanding from multi-level and multi-view images. More precisely, the OmniCity contains multi-view satellite images as well as street-level panorama and mono-view images, constituting over 100K pixel-wise annotated images that are well-aligned and collected from 25K geo-locations in New York City. To alleviate the substantial pixel-wise annotation efforts, we propose an efficient street-view image annotation pipeline that leverages the existing label maps of satellite view and the transformation relations between different views (satellite, panorama, and mono-view).  
With the new OmniCity dataset, we provide benchmarks for a variety of tasks including building footprint extraction, height estimation, and building plane/instance/fine-grained segmentation. Compared with the existing multi-level and multi-view benchmarks, OmniCity contains a larger number of images with richer annotation types and more views, provides more benchmark results of state-of-the-art models, and introduces a novel task for fine-grained building instance segmentation on street-level panorama images. Moreover, OmniCity provides new problem settings for existing tasks, such as cross-view image matching, synthesis, segmentation, detection, etc., and facilitates the developing of new methods for large-scale city understanding, reconstruction, and simulation. The OmniCity dataset as well as the benchmarks will be available at
\url{https://city-super.github.io/omnicity}.

\end{abstract}

\begin{IEEEkeywords}
Datasets, Multi-view images, Street-level images, Remote sensing, Instance segmentation, Height estimation
\end{IEEEkeywords}}

\maketitle

\IEEEdisplaynontitleabstractindextext

%
\IEEEpeerreviewmaketitle



\section{Introduction}

Owning over a half global population and contributing the most economic growth, the city areas have been recorded and characterized by various data sources including satellite and aerial imagery, street-level imagery, LiDAR data, public maps, crowd-sourced data, etc. Over the past few years, a great number of benchmarks have been proposed towards facilitating different vision tasks in city scenes. Among these various data sources, the street-level imagery captured by fixed cameras has been broadly used in multiple benchmarks for driving scenes, e.g. Cityscapes \cite{cordts2016cityscapes}, Mapillary \cite{neuhold2017mapillary}, ApolloScape \cite{huang2019apolloscape}, nuScenes \cite{caesar2020nuscenes}, KITTI \cite{geiger2012we}, Waymo \cite{sun2020scalability}, etc. The rich visual information provided by street-level imagery also enables complicated visual recognition tasks on specific categories, such as person detection (EuroCity dataset \cite{braun2019eurocity}), vehicle tracking and re-identification (CityFlow dataset \cite{tang2019cityflow}), fine-grained land use classification \cite{zhu2019fine}, and identifying their business entity information of buildings \cite{zhang2020building}.

Nevertheless, producing pixel-wise annotations for street-level imagery is a challenging task requiring substantial human efforts, which results in the small image quantity and limited view types of existing street-view datasets. Especially for the street-view panorama datasets, such as TorontoCity \cite{wang2017torontocity}, PASS \cite{yang2019pass} and WildPASS \cite{yang2021context}, the pixel-wise annotated panorama images have a very limited quantity that ranges from $400$ to $520$. 
Regarding the annotation categories and levels, existing street-level datasets mostly provide instance-level annotations for dynamic object categories in driving scenes (i.e. person, vehicles, bicycles, etc.). 
As a vital component for city understanding, the static objects such as buildings take up a larger proportion of cities and remain a higher consistency across the satellite and ground-level images compared with the dynamic objects.
However, existing street-level datasets either provide pixel-wise building annotations without fine-grained semantic labels (such as HoliCity \cite{zhou2020holicity}, driving-scene datasets \cite{cordts2016cityscapes,caesar2020nuscenes,geiger2012we}, etc.) or provide fine-grained building annotations at only bbox or image level (such as \cite{zhu2019fine,zhang2020building}).

\begin{figure*}
\centering
\includegraphics[width=\textwidth]{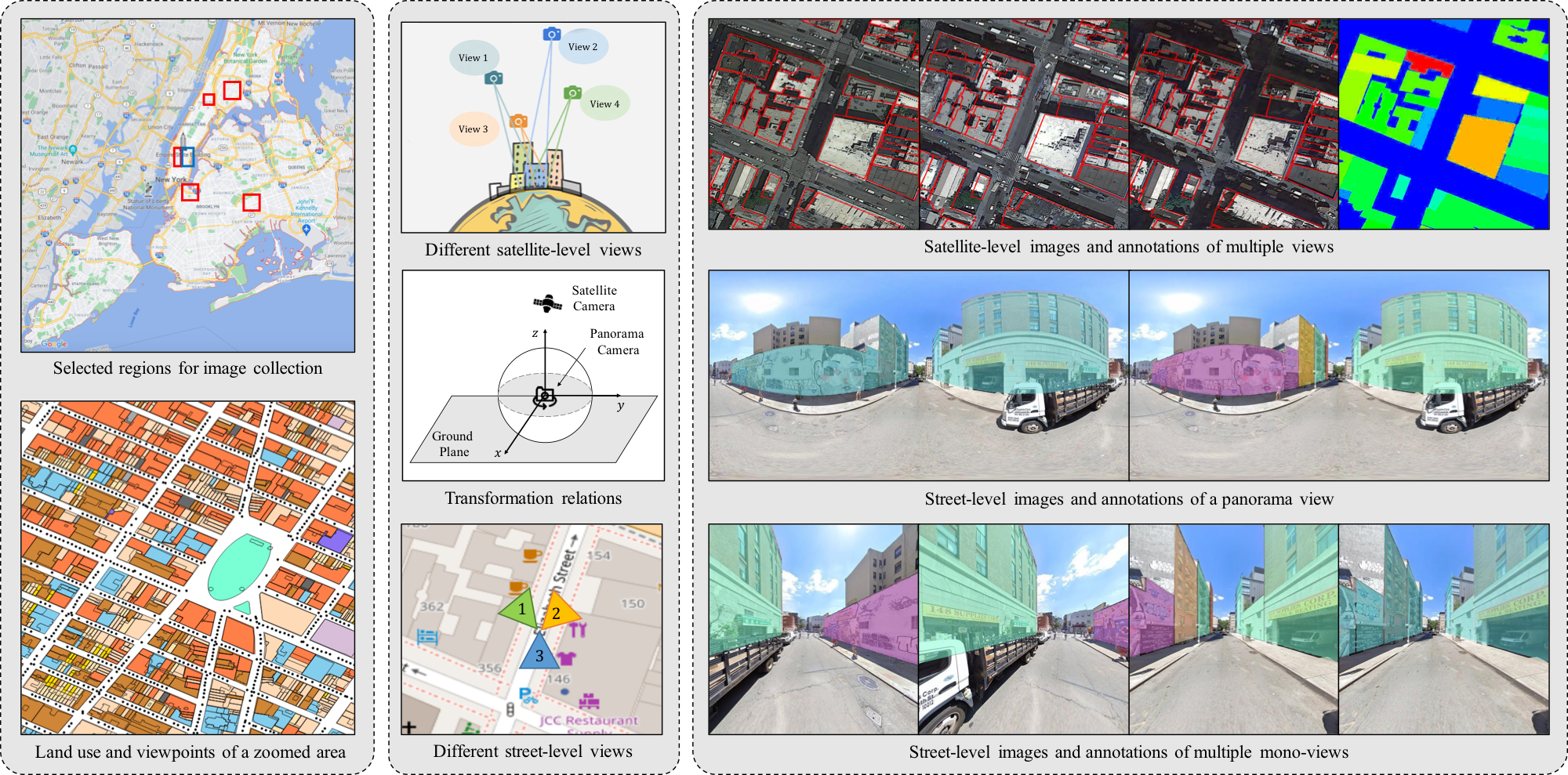}
\caption{An overview of our proposed OmniCity dataset. The satellite and street-level images of our dataset are collected in the six selected regions. Taking a zoomed region as an example, the viewpoint geo-locations represented by black dots are distributed with a step distance of 65 meters along the streets, and different land lot colors indicate different land use categories. The satellite-level annotations include building footprint and heights on images of multiple views. The street-level annotations include plane/instance/fine-grained segmentation on the panorama image and its corresponding mono-view images of different FoVs. Note that all the images in the right part correspond to the same geo-location, and the intrinsic transformation relation between the satellite view and the street-level panorama view is demonstrated in the middle part.}
\label{figure-overview}
\end{figure*}

The remote sensing data, such as the satellite and aerial imagery, has been extensively explored in large-scale city understanding studies for land cover and land use mapping \cite{li2020integrating}, building footprint extraction \cite{li2021joint,wei2019toward,bischke2019multi}, building height estimation \cite{2020Learning,li20213d}, road extraction \cite{liang2019convolutional}, etc.
Compared with the street-level images, the remote sensing imagery from the satellite or aerial view usually contains less visual information. Thus it is extremely difficult to conduct fine-grained land use segmentation, building function recognition, urban zoning segmentation and other complicated tasks merely based on the remote sensing imagery. 
On the other hand, unlike the sparsely-distributed street-level images, the remote sensing images have a dense spatial distribution and a worldwide coverage, which are well aligned with the open GIS maps and government datasets at pixel level.
These existing maps and datasets contain a variety of satellite-level annotations for buildings (such as the footprint, land use, height, year built), roads (category and line coordinates), and other geographical objects, providing new opportunities and perspectives for promoting novel city understanding datasets and tasks.

In this work, as illustrated in Figure~\ref{figure-overview}, we construct an omnipotent city dataset unifying data sources from both satellite and street views, linked by geo-locations and urban planning data. Unlike existing city datasets that only support a limited number of tasks, OmniCity dataset incorporate rich geometric annotations and semantic meta data for each image, where multiple tasks can be conducted on.
To leverage the existing satellite-level annotations and the rich visual context from the street-level imagery, we propose a novel annotation pipeline for producing diverse street-level annotations with fewer human labeling efforts. Specifically, the annotators consider both auxiliary information (transformed from the satellite-level annotations) and building appearances (e.g. texture discrepancy, doors, etc.) to efficiently annotate each building on panorama images, which are further converted to multiple mono-view images and annotations automatically via a view transformation method. 
Based on our proposed annotation pipeline, we built OmniCity, a dataset that contains over 100K annotated images collected from 25K geo-locations in New York City. Compared with existing datasets, OmniCity provides richer annotation types for well-aligned satellite and street-level images captured from multiple views of each geo-location, enabling more omnipotent city understanding via a diversity of tasks.

We conduct experiments on our OmniCity dataset for a variety of tasks, including building footprint extraction and height estimation tasks on satellite images, as well as the fine-grained/instance/plane segmentation of buildings on street-level panorama and mono-view images. To the best of our knowledge, this is the first work that involves fine-grained building instance segmentation task on street-level panorama images. Compared with current multi-level and multi-view benchmarks, OmniCity provides baseline experimental results obtained from more state-of-the-art models and additional semantic-related tasks. 
Moreover, we conduct results analysis from various perspectives, including the impact of view angles and types on model performance, the performance of different methods for satellite and street-level tasks, and the limitations of current methods for different tasks on panorama images.
We also analyze the potential of OmniCity for promoting novel problem settings, tasks and algorithms with multi-level or cross-view imagery.

\linespread{1.2}
\begin{table*}[]
\centering
\caption{A comparison of our proposed dataset and existing city-related datasets. The \# Images column represents the number of annotated images. The street view column shows whether the dataset contains no / mono-view (mono) / panorama (pano) street-level images. The satellite view column shows whether the dataset contains no / single / multiple satellite images. The annotation level column indicates which level of tasks the dataset is designed for, i.e., semantic segmentation, object detection (bbox), instance segmentation, plane segmentation, and image classification. The last two columns indicate whether the dataset contains fine-grained land use or height labels.}
\label{table-datasets}
\resizebox{\textwidth}{!}{%
\begin{tabular}{|c|c|c|c|c|c|c|c|}
\hline
Type                             & Dataset         & \# Images & Street view  & Satellite view & Annotation level & Land use & Height  \\ \hline
\multirow{6}{*}{Street-level}    & KITTI  \cite{geiger2012we}         & 15,000    & mono         & no             & semantic         & no           & no          \\ \cline{2-8} 
                                 & Cityscapes \cite{cordts2016cityscapes}     & 25,000    & mono         & no             & semantic         & no           & no          \\ \cline{2-8} 
                                 & EuroCity \cite{braun2019eurocity}       & 47,300    & mono         & no             & bbox             & no           & no          \\ \cline{2-8} 
                                 & WildPASS \cite{yang2021context}   & 500       & pano \& mono &  no             & semantic         & no           & no          \\ \cline{2-8} 
                                 & PASS  \cite{yang2019pass}          & 400       & pano \& mono & no             & semantic         &no           &no          \\ \cline{2-8} 
                                 & HoliCity \cite{zhou2020holicity}       & 6,300     & pano \& mono & no             & instance/plane   &no           &no          \\ \hline
\multirow{4}{*}{Satellite-level} & SkyScapes \cite{azimi2019skyscapes}       & 8,820     & no           & single         & semantic         &no           & no          \\ \cline{2-8} 
                                 & SpaceNet MVOI \cite{0SpaceNet}  & 60,000    & no           & multiple       & instance         & no           & no          \\ \cline{2-8} 
                                 & Christie et al. \cite{2020Learning} & 11,000    & no           & single         & semantic         & no           & yes         \\ \cline{2-8} 
                                 & Li et al. \cite{li20213d}      & 3,300     & no           & single         & instance         & no           & yes         \\ \hline
\multirow{3}{*}{Cross-level}     & TorontoCity \cite{wang2017torontocity}    & unknown   & pano \& mono & single         & instance         & no           & yes         \\ \cline{2-8} 
                                 & Wojna et al. \cite{wojna2021holistic}   & 49,426    & mono         & single         & image            & yes          & no          \\ \cline{2-8} 
                                 & Ours            & \textbf{108,600}    & \textbf{pano \& mono} & \textbf{multiple}       & \textbf{instance/plane}   & \textbf{yes}          & \textbf{yes}         \\ \hline
\end{tabular}%
}
\end{table*}

Our main contributions are summarized as follows:

\begin{itemize}

\item We propose a novel pipeline for efficiently producing diverse pixel-wise annotations on street-level panorama and mono-view images.

\item We build the OmniCity dataset, which contains well-aligned satellite and street-level images with a larger quantity, richer annotations and more views compared with existing datasets.

\item We provide a series of baseline experiments for 2D and 3D tasks on multiple data sources, and a comprehensive analysis of many aspects including the impact of views, limitations of current methods, etc.

\item We discuss the potential of our proposed OmniCity for facilitating new problem settings, methods and tasks for large-scale city understanding, reconstruction, and simulation.

\end{itemize}

\section{Related work}

\subsection{Datasets and methods for street-level tasks}

As shown in Table \ref{table-datasets}, many street-level datasets have been proposed over the past few years, of which a large proportion is aiming at the visual tasks in driving scene (e.g. object detection, object tracking, re-identification, semantic segmentation, etc.). 
As a pioneer in this field, the KITTI dataset \cite{geiger2012we} provides dense point clouds and stereo images to support multiple tasks such as 2D and 3D object detection, object tracking, etc. For 2D semantic segmentation or detection tasks, the Cityscapes \cite{cordts2016cityscapes} and Mapillary \cite{neuhold2017mapillary} datasets contain thousands of pixel-wise annotated images for semantic understanding of street Scenes, while the BDD100K dataset \cite{yu2020bdd100k} is comprised of both 100 thousand images with bbox-level annotations and 10 thousand images with pixel-wise annotations. Similar to KITTI \cite{geiger2012we}, two recent released datasets named ApolloScape \cite{huang2019apolloscape} and nuScenes \cite{caesar2020nuscenes} provide both 2D and 3D annotations for multiple visual tasks such as detection, segmentation, stereo, localization, tracking, etc. The Waymo dataset \cite{sun2020scalability}, by contrast, provides 2D and 3D bounding boxes for detection and tracking tasks instead of semantic labeling tasks. Several other studies propose street-level datasets for vision tasks of a specific object category, such as the EuroCity Persons dataset \cite{braun2019eurocity} for object detection and the CityFlow dataset \cite{tang2019cityflow} for vehicle tracking and re-Identification in multi-target and multi-camera conditions. 

The above datasets are manually annotated by human annotators. To reduce the expensive human efforts for creating pixel-wise annotations, Richter et al. \cite{richter2016playing} proposed a novel approach to rapidly creating pixel-accurate semantic label maps for images extracted from modern computer games, producing the GTA5 dataset with around 25,000 pixel-wise labeled images of a single street view. 
In addition to the above street-level datasets containing only mono-view images, some studies propose new datasets or methods for semantic segmentation from panorama images, such as TorontoCity \cite{wang2017torontocity}, PASS \cite{yang2019pass}, and WildPASS \cite{yang2021context}. However, each of the above dataset requires expensive human efforts for creating the panorama annotations, which contains only 400 to 520 annotated panorama images in total.

Besides the street-level datasets collected for driving scenes, several recent studies target at the recognition tasks for static object categories such as buildings and land use. For instance, Zhu et al. \cite{zhu2019fine} proposed a framework for fine-grained land use classification at the city scale using ground-level images. Although producing a pixel-wise land use map from the satellite view, the proposed dataset and framework is designed for the image-wise classification task on street-level images. In Zhang et al. \cite{zhang2020building}, an end-to-end building recognition system is proposed for detecting buildings and identifying their business entity information from street-side images, in which the building annotations are provided at bounding box level. 
HoliCity \cite{zhou2020holicity} is a recent released dataset with various annotations of holistic 3D structures (such as line, wireframe, surface normal, depth, etc.) that are generated from CAD models. Although HoliCity provides both mono-view and panorama images with various view angles, the study only conduct experiments on mono-view images. Moreover, no semantic categories of buildings are provided in HoliCity dataset.

In summary, existing street-level datasets still have the following limitations. Regarding the image quantity and view types, most existing datasets require substantial annotation efforts and contain only a limited number of annotated images collected from a mono view. The image quantity of panorama datasets is even several orders of magnitude smaller than the datasets containing only mono-view images.
Regarding the annotation categories and levels, existing datasets mainly focus on dynamic object categories (person, vehicles, bicycles, etc.) or driving-related categories (traffic signs, lanes, etc) and lack in the fine-grained types of each object category. Moreover, the datasets for static objects such as buildings are either lacking in pixel-level annotations or fine-grained building types.
Our OmniCity, on the contrary, contains over 100K satellite and street-level images of multiple views, as well as the pixel-wise and fine-grained annotations, which facilitates new problem settings and tasks for large-scale city understanding, reconstruction and simulation compared with existing datasets.

\subsection{Datasets and methods for satellite-level tasks}

As a data source with a long time series and a large coverage, the satellite imagery has been broadly explored for large-scale city understanding. Unlike the street-level datasets that usually require manual annotations, there are already many existing pixel-wise label maps that are well aligned with the satellite imagery. The OpenStreetMap (OSM)\footnote{\url{http://wiki.openstreetmap.org}.} is one of the most popular map data among various global maps (e.g. Google, Bing, Yahoo maps, etc.), which contains publicly available and collaboratively editable annotations of many types (building footprints, building heights, roads, land use, etc.) of world wide. In addition to the global maps, many public datasets provide rich information at a local scale. For example, the PLUTO \footnote{\url{https://data.cityofnewyork.us/City-Government/Primary-Land-Use-Tax-Lot-Output-PLUTO-/64uk-42ks}.} dataset provided by New York government contains the block and lot information of the whole New York City, and each lot is associated with the land use, year built, number of floors, and other useful information. The Microsoft US building footprint dataset\footnote{\url{https://github.com/microsoft/USBuildingFootprints}.} contains over a hundred million computer-generated building footprints in all 50 US states, with better or similar metrics compared to OSM building metrics against the labels. Some challenge datasets provide manually labeled building footprints as well as the corresponding single-view, multi-view, or multi-temperal satellite images, such as the DeepGlobe 2018 \cite{2018DeepGlobe}, the SpaceNet Multi-View Overhead Imagery Dataset (MVOI) \cite{0SpaceNet}, and the Multi-Temporal Urban Development SpaceNet Dataset \cite{van2021multi}. Moreover, some datasets provide fine-grained semantic labels with tens of categories in addition to the building footprints (e.g. SkyScapes \cite{azimi2019skyscapes}), or provide the building height annotations for height estimation and 3D reconstruction tasks \cite{2019Semantic,2020Learning,li20213d}.

In summary, the existing maps and datasets have provided a large amount of semantic and geometric information that is well aligned with the satellite imagery. In this work, we leverage these rich annotations from satellite view and the transformation relations between different views (i.e., satellite to panorama views, and panorama to mono views) to produce auxiliary information for street-level image annotation. Compared with the existing datasets, our OmniCity significantly reduces the human labeling efforts required for the large-scale street-level datasets, and provides more annotation types to enable omnipotent city understanding via multiple tasks and views.

\subsection{Datasets and methods for multi-level tasks}

The ground-level imagery usually contains rich visual context that is not visible from the satellite or aerial imagery (e.g. the building facade, the side of vegetations, etc.), while the spatial distribution of ground-level images is often sparse and unbalanced in different areas. The satellite and aerial imagery, on the contrary, has a much denser spatial distribution at a global scale, but the visual context is too limited to accurately accomplish complicated tasks such as building function recognition, urban zoning segmentation, etc. Considering the complementary characteristics of the two data sources, many datasets and methods have been proposed towards visual recognition, image matching, image synthesis, or multiple tasks based on cross-view images.

For the cross-view visual recognition task, many studies tried to integrate the two data types and leverage their characteristics to produce a dense map for building functions, building ages, land use, tree species, or other visual recognition tasks \cite{wegner2016cataloging,workman2017unified,feng2018urban,koh2020sideinfnet,kang2018building,srivastava2019understanding,suel2021multimodal}. These studies require both satellite and ground-level images as the input.
By contrast, the cross-view image matching (ground-to-aerial image geo-localization) and image synthesis tasks take only one data type as the input, which have been extensively studied over the past few years. As a pioneer work for cross-view image matching, Lin et al. \cite{lin2013cross} proposed a cross-view feature translation approach to enable localizing a query image even without corresponding ground-level images in the database. Later studies further improved the image geo-localization performance \cite{tian2017cross,hu2018cvm,zhu2021vigor}, proposed novel cross-view localization datasets \cite{regmi2019bridging}, or introducing novel tasks such as estimating the orientation information between cross-view images \cite{vo2016localizing,shi2020looking,zhu2021revisiting}.
For cross-view image synthesis, a consistent ground-view image is generated from an input satellite image \cite{tang2019multi,toker2021coming,cai2019ground,shi2019spatial,li2021sat2vid}. Early studies aimed at predicting the ground-level image using only satellite imagery as the supervision \cite{zhai2017predicting}, while recent methods leveraged both satellite imagery and its corresponding semantic segmentation map as the supervision \cite{regmi2018cross}, or introduced extra geometric information (e.g. the height map) and a geo-transformation method to produce more realistic ground-level imagery \cite{lu2020geometry,shi2022geometry}.

\begin{figure*}
\centering
\includegraphics[width=\textwidth]{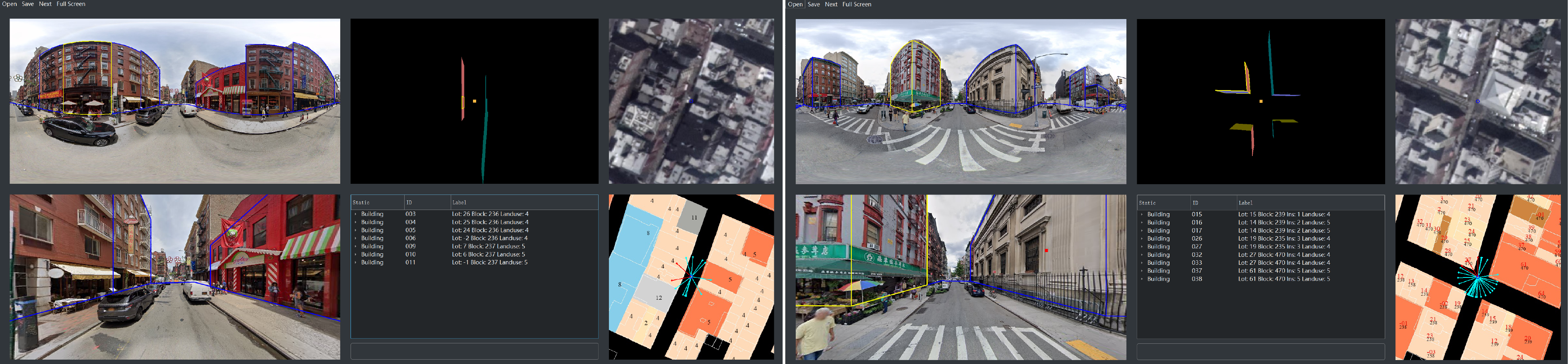}
\caption{Annotation GUI of our proposed OmniCity dataset. The left image shows an example of the panorama image collected in one road scene, with the label map (in the bottom-left window) displayed in land use mode. The right image shows an example of the panorama image collected in crossroads scene, with the label map displayed in block-lot mode.}
\label{fig:long}
\label{figure-annotation-tool}
\end{figure*}

Different from the above studies that are designed for one specific task, several studies provide a variety of tasks for satellite or street-level images.
The TorontoCity \cite{wang2017torontocity} contains a wide range of annotations including building height estimation, building instance segmentation, building footprint segmentation, road segmentation, etc., which are conducted on either satellite or street-level images. However, the study only adopt FCN-based architectures and some CNN classification networks as the baseline models for all experiments, and the dataset is still not available for public use. 
In a recent study \cite{wojna2021holistic}, a holistic multi-view building analysis dataset is designed for multiple recognition tasks regarding facade material, number of floors, occupancy type, roof geometry, roof pitch, and construction type. However, the dataset contains only image-level annotations for street-level classification tasks. The semantic categories (i.e. facade material, occupancy type, construction type, etc.) of each street-level image are decided by human annotators, resulting in more challenges and subjectivity to the annotation process compared with our work.

In summary, for the existing cross-view studies designed for one specific task (i.e., visual recognition, image matching and image synthesis), the street-level images used in these studies are not annotated and only supplement extra feature to the satellite-level feature maps.
Our OmniCity dataset, on the contrary, provides rich annotation types that are not contained in existing datasets, e.g. the pixel-wise building instances and fine-grained categories on street-level images, which might promote new methods to explore and leverage these new annotations to further improve the performance of these tasks. 
In addition, different from the existing studies designed for multiple tasks, OmniCity provides the baseline experimental results of each task using more state-of-the-art models. With our proposed efficient annotation pipeline, it also provides additional fine-grained pixel-wise annotations and benchmark results compared with \cite{wang2017torontocity} and \cite{wojna2021holistic}.

\section{Datasets}

In this work, we aim at building a dataset for omnipotent city understanding from multi-level and multi-view images. Our proposed OmniCity dataset contains 108,600 images of multiple views, which are collected from 25,000 geo-locations in New York City.
Compared with existing datasets, OmniCity requires much fewer human efforts for street-level image annotation, contains more diverse annotation types for both 2D and 3D tasks, provides richer building semantics at instance segmentation level, and possesses higher scalability for new annotation type supplement and expansion to other cities.
The details of data collection, annotation, and statistics will be introduced as follows.

\subsection{Data collection}

As shown in Figure \ref{figure-overview}, our OmniCity dataset is collected from six selected regions of New York. For the street-level tasks, we download the panorama images in the six selected regions using google street view download 360, with a step distance of 65 meters. 
The five regions for collecting training and validation samples are denoted by red color, which are randomly divided using a ratio of 3:1. The region for collecting test samples is denoted by blue color, which is distributed separately with the training and validation regions.
The panorama image collection time is between 2019 and 2020. We save the geographic coordinates (longitude and latitude), collection time (year and month), panorama id, north rotation, and zoom level for each panorama image. The mono-view images are automatically generated from the panorama images, which will be introduced in Section \ref{sec-annotation}.
For each panorama site, we download its corresponding google earth images of three acquisition dates according to the geographic coordinates, constituting three groups of datasets for our satellite-level tasks. As shown in Figure \ref{figure-overview}(c), the satellite images in the three groups of datasets have three types of off-nadir view angles, which are denoted by small, medium and high in the following sections.
All satellite images patch is in 512 $\times$ 512 pixels, with a spatial resolution of around 0.3 meters (Level 19).

For the annotation data sources, we collect the meta information from PLUTO (provided by New York government) and OpenStreetMap (OSM). The New York City is hierarchically formed by blocks, lots, and buildings. Each building can be identified by a specific block-lot id. In PLUTO, each lot (building) is associated with rich information, e.g. land use, year built, number of floors, etc. The OpenStreetMap data contains complete building footprint and height information in New York area, while lacking in land use information for many building instances. Considering the characteristics of the two data sources, we align the land use attribute (from PLUTO) with the building footprint and height (from OSM) using the geographical coordinates. Overall, each building is assigned with a block-lot id, a land use category, a height value, and the geographical coordinates of a footprint polygon, which will be used as the reference label maps for panorama image annotation.

\begin{figure*}
\centering
\includegraphics[width=\textwidth]{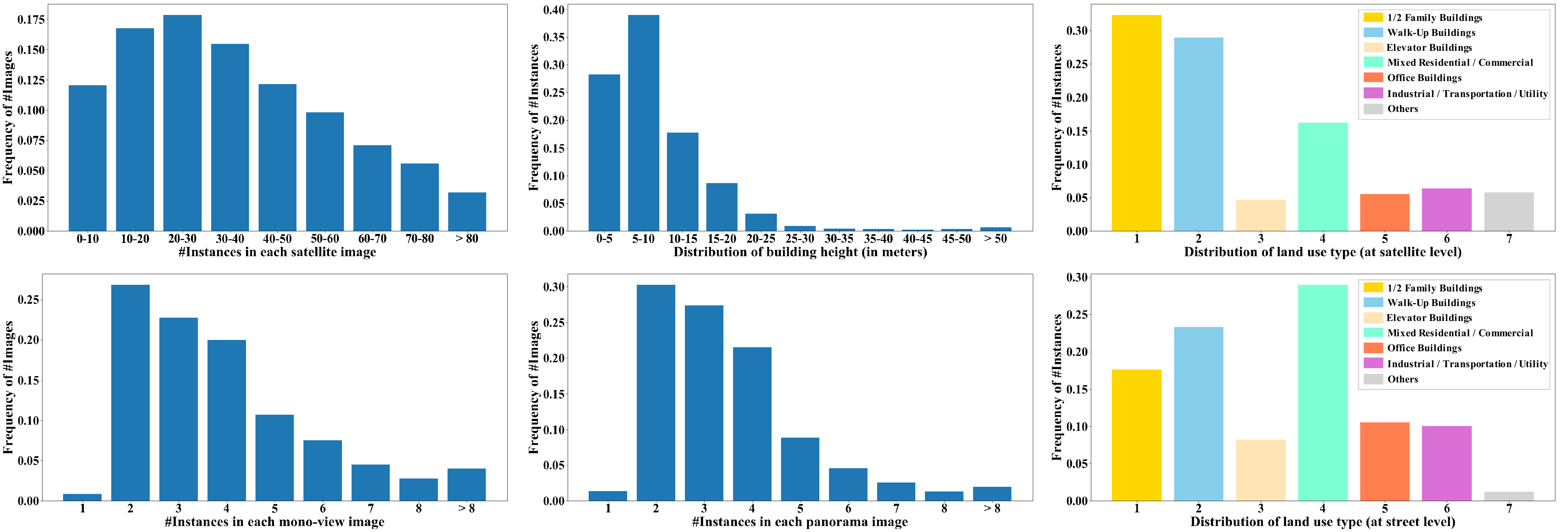}
\caption{Statistics of the proposed dataset, in terms of the instance quantity in each satellite/mono-view/panorama image, the distribution of building height, and the distribution of land use type at satellite/street level.}
\label{figure-statistics}
\end{figure*}

\subsection{Data annotation}\label{sec-annotation}

In this section, we introduce the annotation pipelines for producing the various types of labels for satellite-level and street-level tasks. 
For satellite-level tasks, the labels are already well-aligned with the meta information from the New York government and the OpenStreetMap. Each building footprint on a satellite image is assigned with a block-lot id, a land use category, a height value, and the pixel coordinates of the footprint polygon, which are converted into the COCO dataset format \cite{lin2014microsoft} for instance segmentation tasks. We further convert the building height annotations into pixel-wise labels for height estimation tasks, i.e., the pixels in footprint regions are set as the instance-wise height values and the pixels in other regions are set as zero. 
For the street-level tasks, we ask human annotators to label the panorama images using our proposed annotation tool, of which the GUI can be found in Figure \ref{figure-annotation-tool}. The labels of the mono-view images are automatically converted from those of panorama images via a transformation method. The procedures for producing the two types of street-level annotations will be introduced as follows.

The annotation pipeline of panorama images includes four stages, i.e. image selection, segmentation annotation, attribute assignment, and quality assessment. The panorama image annotation tool proposed in our study is developed based on an existing tool for labeling panorama images of indoor scene\footnote{\url{https://github.com/SunDaDenny/PanoAnnotator}.}. We supplement many new functions to enable the labeling of OmniCity dataset. As shown in Figure \ref{figure-annotation-tool}, the GUI contains six sub-windows, of which each window (from left to right and top to bottom) shows: (1) the panorama image with the current annotation; (2) the current annotation result in 3D space; (3) the corresponding satellite image; (4) the mono-view image with the current annotation; (5) the attributes of the labeled building instances; (6) the visualized label map of the corresponding satellite image. 

\noindent \textbf {Image selection:} The annotator is required to decide whether a panorama image is essential to be labeled. The panorama image that meets any of the following conditions will be regarded as an invalid image and skipped, i.e., building areas take up only a small percentage of the whole image, buildings only locate in the side of the panorama image, and buildings are seriously sheltered by trees, vehicles or mosaics; Otherwise, the panorama image will be selected as a valid image and labeled in the following stages. 

\noindent \textbf {Segmentation annotation:} The annotator is first required to drag the floor line to fit the bottom boundary of all buildings. Next, the annotator needs to add the split line and adjust the top line to fit the roof boundary for each building plane. In the bottom-right sub-window, we provide auxiliary information indicating the approximate locations of the split lines, which is generated by transforming the building footprint split lines in the satellite view to panorama view using a geo-transformation method \cite{lu2020geometry}. The annotators should consider both auxiliary information and building appearance (e.g. texture discrepancy, doors, etc.) to decide the accurate location of each split line. 
Note that, as the urban planning data and footprint data are likely to be out-of-date and may not match with the timestamps the street-view panoramas are collected, it is necessary for human annotators to balance multiple considerations to make proper adjustments to the building boundaries.

\noindent \textbf {Attribute assignment:} The annotator needs to add the attributes (instance ID, block-lot id and landx use type) for each building plane labeled in the segmentation annotation stage, which are demonstrated in the bottom-middle sub-window. The building planes that belong to the same building instance will be set as the same instance ID (in the crossroads scene); Otherwise, the plane will be set as a specific instance ID successively. When a building instance is selected by the annotator (the yellow one in the panorama image), the surrounding auxiliary lines of its corresponding footprint in the bottom-right sub-window will turn red. Then the annotators assign the lot-block id and the land use type according to the numbers shown in the bottom-right sub-window, which can be switched between the land use mode and the block-lot mode. 

\noindent \textbf {Quality assessment:} We ask another group of experienced annotators to check the quality of all annotated panorama images. These annotators remove the invalid panorama images that are not deleted in the image selection stage, and those with wrong annotations resulting from omitting attribute assignment operations, adding inaccurate split lines, labeling only a portion of planes, etc. The remaining valid images and labels constitute the final panorama dataset that will used in our experiment.

\begin{figure}[t!]
\centering
\includegraphics[width=\linewidth]{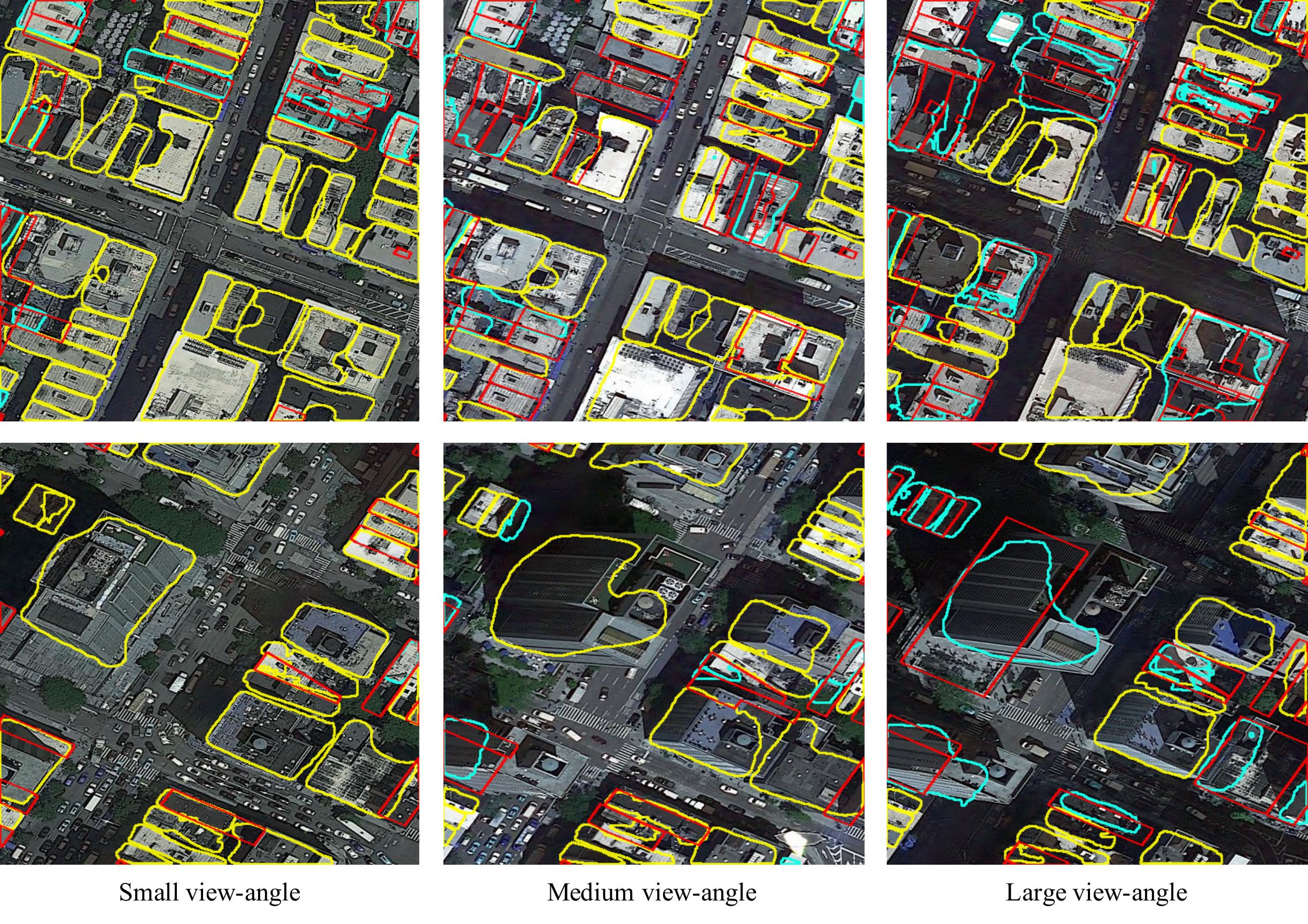}
\caption{Baseline results of building instance segmentation on satellite images. The yellow, cyan, and red polygons denote the TP, FP, and FN.}
\label{fig:long}
\label{figure-satseg-3view}
\vspace{-3mm}
\end{figure}

The mono-view images and the corresponding annotations can be automatically generated from those of panorama images via a transformation method\footnote{\url{https://github.com/sunset1995/py360convert}.}. For each panorama image, we select three views using three x-axis angles (-170, 10 and 170, in the range of [-180, 180]) and a fixed y-axis angle of 0. The size of each mono-view image is set as 512 $\times$ 512 pixels. Then we design a mono-view image selection procedure to filter out the unexpected views using the following three rules: (1) buildings are distributed in only one side of the image; (2) the percentage of building area is smaller than 10\%; (3) the number of building planes is smaller than 2. The remaining valid images and annotations that do not meet any of the three conditions constitute the final mono-view dataset that will used in our experiment.

\subsection{Statistics of the proposed dataset}

The whole dataset contains three sub-datasets (in small, medium and large view angles) for satellite-level tasks and two sub-datasets (in panorama and mono views) for street-level tasks. For satellite-level datasets, each image in the three sub-datasets is collected from the 25,000 geo-locations of the six selected regions, constituting 75,000 satellite image patches in 512 $\times$ 512 pixels in total. For the street-level datasets, the panorama dataset contains 18,000 images in 512 $\times$ 1024 pixels, which are selected from the initial 25,000 panorama images during the image selection and quality assessment phases. Similarly, the mono-view dataset is obtained from the panorama dataset via a view transformation and a mono-view image selection procedure, constituting around 156,00 mono-view images in 512 $\times$ 512 pixels. In summary, the whole OmniCity dataset contains 108,600 annotated images in total.
For all five sub-datasets, the ratio of train/val/test splits is set as 3:1:1.

Figure \ref{figure-statistics} shows the statistics of our OmniCity dataset, in terms of the instance quantity per satellite/mono-view/panorama image, the distribution of building height, and the distribution of land use type at satellite/street level. The instance quantity per satellite image ranges from 1 to 86, which is much higher than those in mono-view and panorama images (1 to 10). This is because the satellite image has a larger geographical coverage compared with the corresponding street-level images with only roadside instances. In OmniCity dataset, the initial land use categories with similar characteristics and low instance quantity are merged into one category, resulting in 7 land use categories in total. Over 70\% instances belong to 1/2 family building, walk-up building and the mixed residential/commercial building categories, while the other four categories (i.e., elevator buildings, office buildings, industrial/transportation/utility buildings and others) take up a relatively small proportion. For the distribution of building height, most buildings are 1 to 25 meters high, and a small percentage of buildings reaches a height of over 50 meters.

\section{Benchmark results}

In this section, we provide a variety of benchmarks for multiple satellite and street-level tasks. The satellite-level tasks in our experiments include building footprint segmentation and height estimation. For both tasks, we conduct experiments on the satellite images with three view angles. For the street-level tasks, we conduct two instance segmentation tasks (i.e., land use and building instance segmentation) on the panorama images, and three instance segmentation tasks (i.e., land use / building instance / plane segmentation) on mono-view images. Please note that these are only preliminary experimental results on OmniCity dataset. More benchmarks of latest models and additional tasks will be continuously updated on OmniCity homepage.

\subsection{Experimental setting}\label{sec-setting}

We use Mask R-CNN \cite{he2017mask} as the baseline method for all segmentation tasks in this section, and provide a results comparison of different instance segmentation methods in section \ref{sec-different-methods}. All methods are implemented base on mmdetection \cite{chen2019mmdetection} with the recommended hyper-parameter settings.
Specifically, we use ResNet-50 \cite{2016Deep} with FPN \cite{lin2017feature} that is pre-trained on the ImageNet \cite{russakovsky2015imagenet} as the backbone for all instance segmentation models. All models are trained on 8 NVIDIA Tesla V100 GPUs for 12 epochs, with a batch size of 16, a learning rate starting from 0.02 and decreasing by a factor of 0.1 from the $8^{th}$ to $11^{th}$ epoch, and the stochastic gradient descent (SGD) optimizer with a weight decay of 0.0001 and a momentum of 0.9.


For the satellite-level height estimation task, we evaluate and analyze the performance of two widely-used monocular depth estimation methods on the satellite images of three view angles, i.e. Structure-Aware Residual Pyramid Network (SARPN) \cite{chen2019structure} and Deep Ordinal Regression Network (DORN) \cite{fu2018deep}. Both SARPN \cite{chen2019structure} and DORN \cite{fu2018deep} are implemented using PyTorch \cite{paszke2019pytorch}. For SARPN, the SENet encoder is initialized by a model pretrained on ImageNet \cite{russakovsky2015imagenet} and the other layers are randomly initialized. The model is trained on 4 NVIDIA Tesla V100 GPUs for 20 epochs, with a batch size of 8, a learning rate starting from $10^{-4}$ and reduced to 10\% every 5 epochs, and the Adam optimizer with $\beta_1$ = 0.9, $\beta_2$ = 0.999, and a weight decay of $10^{-4}$. The DORN model is trained on 1 NVIDIA Tesla V100 GPUs for 20 epochs, with a batch size of 4, a base learning rate of $10^{-4}$ and the power of 0.9, using SGD optimizer with a weight decay of 0.0005 and a momentum of 0.9.

\begin{table}[t!]
\centering
\caption{Baseline results of instance segmentation for satellite images with different view angles, in terms of COCO and SpaceNet evaluation metrics.}
\label{table-satseg-3view}
\resizebox{\linewidth}{!}{%
\begin{tabular}{|c|c|c|c|c|c|c|c|c|c|}
\hline
\multicolumn{1}{|c|}{\multirow{2}{*}{View}} & \multicolumn{6}{c|}{Metrics of various thresholds}                                             & \multicolumn{3}{l|}{threshold = 0.5} \\ \cline{2-10} 
\multicolumn{1}{|l|}{}                            & $AP$            & $AP_{50}$        & $AP_{75}$        & $AP_S$         & $AP_M$         & $AP_L$         & P      & R        & F1           \\ \hline
Small                                             & \textbf{29.7} & \textbf{66.0} & \textbf{23.5} & \textbf{15.9} & \textbf{33.9} & \textbf{36.7} & \textbf{76.9}  & \textbf{66.3}  & \textbf{71.2} \\ \hline
Medium                                            & 23.7          & 56.6          & 16.1          & 11.5          & 27.2          & 30.3          & 73.9           & 55.0           & 63.1          \\ \hline
Large                                             & 18.9          & 51.4          & 9.6           & 9.1           & 21.5          & 25.3          & 70.7           & 51.7           & 59.7          \\ \hline
\end{tabular}%
}
\end{table}

\begin{figure*}[t!]
\centering
\includegraphics[width=\textwidth]{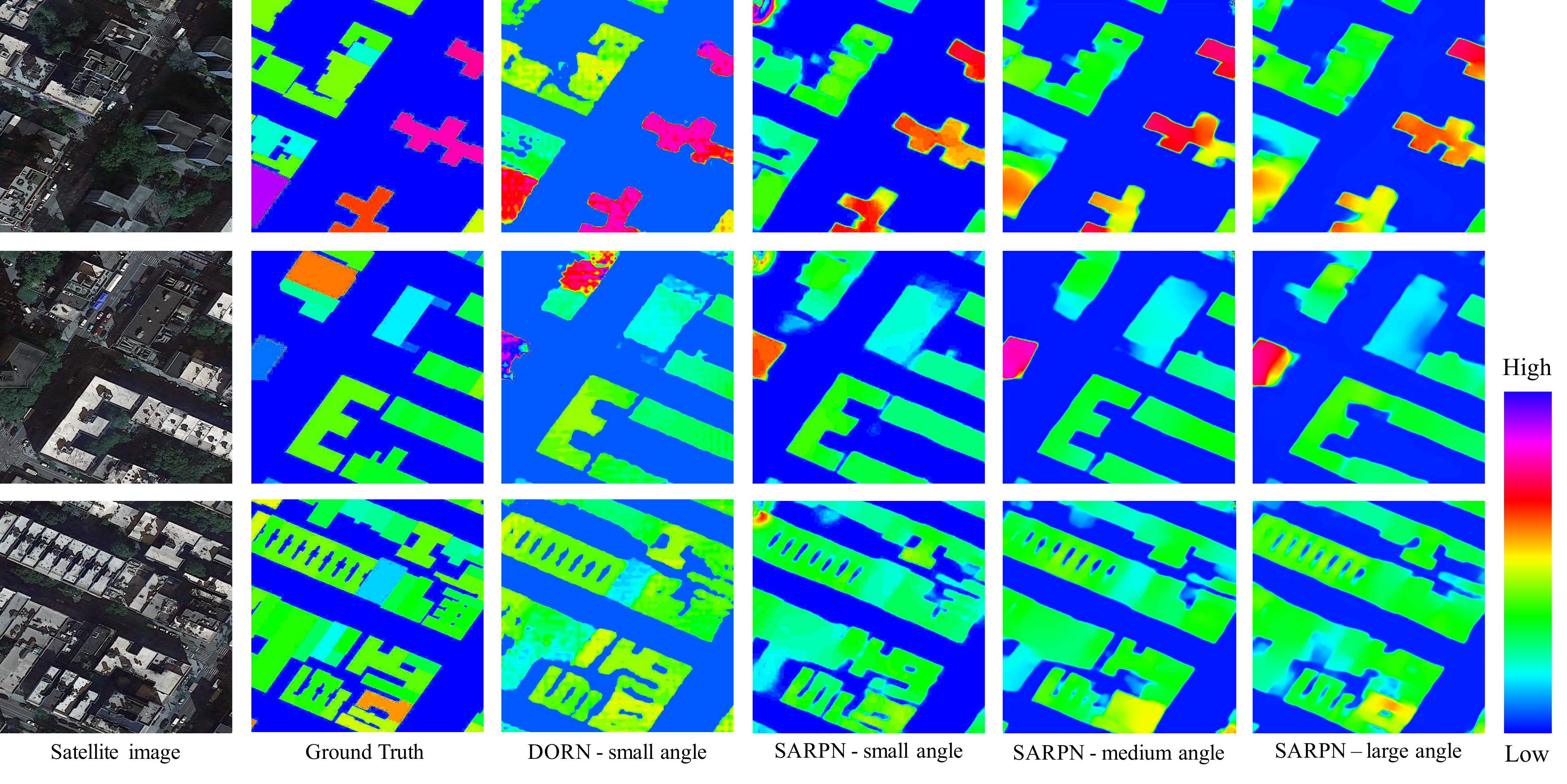}
\caption{Height estimation results obtained from DORN \cite{fu2018deep} and SARPN \cite{chen2019structure} for satellite images with different view angles. Different colors represent the height estimation results of different values.}
\label{fig:long}
\label{figure-height}
\end{figure*}

\subsection{Baselines for satellite-level tasks}

\subsubsection{Building instance segmentation}

We evaluate and analyze the building footprint segmentation performance using both COCO \cite{lin2014microsoft} and SpaceNet \cite{2018DeepGlobe} evaluation metrics. Specifically, we report AP, $AP_{50}$, $AP_{75}$, $AP_{S}$, $AP_{M}$, and $AP_{L}$ for COCO metrics, and the precision, recall and F1-score under a fixed IoU threshold of 0.5 (denoted by P, R, and F1) for SpaceNet metrics. Table \ref{table-satseg-3view} shows the baseline results of the satellite-level instance segmentation task. The footprint segmentation performance is the best for satellite images with a small view angle, and deteriorates seriously when the view angle gets larger. Moreover, for all three cases, the AP is the highest for buildings with large areas and the lowest for small buildings. Figure \ref{figure-satseg-3view} provides a qualitative comparison of the footprint segmentation results on satellite images of three types of view angle, in which the yellow, cyan and red polygons denote the true positive (TP), false positive (FP) and false negative (FN) buildings. The large view angle results in great difficulties for extracting accurate footprint boundaries, due to the partial invisibility of building footprint, the serious shadow effects, etc.

\subsubsection{Building height estimation}

The height estimation performance is evaluated in terms of the mean absolute error (denoted by MAE), mean square error (denoted by MSE), and root mean square error (denoted by RMSE), which are commonly used metrics for depth estimation. All metrics are measured in meters at pixel level. Table \ref{table-sat-height-3view} lists the height estimation results obtained from SARPN \cite{chen2019structure} and DORN \cite{fu2018deep} for satellite images with different view angles. Figure \ref{figure-height} shows the visualized height estimation results and ground truth. Results demonstrate that DORN obtains better quantitative results and more accurate height values compared with SARPN for all three cases, reducing the MSE by 3.47, 1.51, and 1.92, respectively. However, the visualized height estimation results of SARPN have more accurate footprint boundaries compared with DORN that produces more noises. Both methods achieve the best height estimation performance for satellite images with a medium view angle compared with the other two cases. For satellite images with a small view angle, the footprint and roof have more overlaps and provide less information for height estimation. On the other hand, the the shadow and parallax effect become more serious with the increase of off-nadir view angle. The above aspects result in challenges for the accurate estimation of building height for satellite images with small and large view angles.

\begin{table}[t!]
\centering
\caption{Baseline results of height estimation for satellite images with different view angles, in terms of MAE, MSE and RMSE evaluation metrics.}
\label{table-sat-height-3view}
\resizebox{\linewidth}{!}{%
\begin{tabular}{|c|c|c|c|c|c|c|}
\hline
\multirow{2}{*}{View} & \multicolumn{3}{c|}{SARPN \cite{chen2019structure}}                        & \multicolumn{3}{c|}{DORN \cite{fu2018deep}}                         \\ \cline{2-7} 
                            & MAE            & MSE             & RMSE           & MAE            & MSE             & RMSE           \\ \hline
Small                       & 16.18          & 870.34          & 29.50          & 12.71          & 670.52          & 25.89          \\ \hline
Medium                      & \textbf{13.75} & \textbf{694.17} & \textbf{26.35} & \textbf{12.24} & \textbf{628.06} & \textbf{25.06} \\ \hline
Large                       & 15.32          & 823.01          & 28.69          & 13.40          & 730.67          & 27.03          \\ \hline
\end{tabular}%
}
\end{table}

\begin{figure*}
\centering
\includegraphics[width=\textwidth]{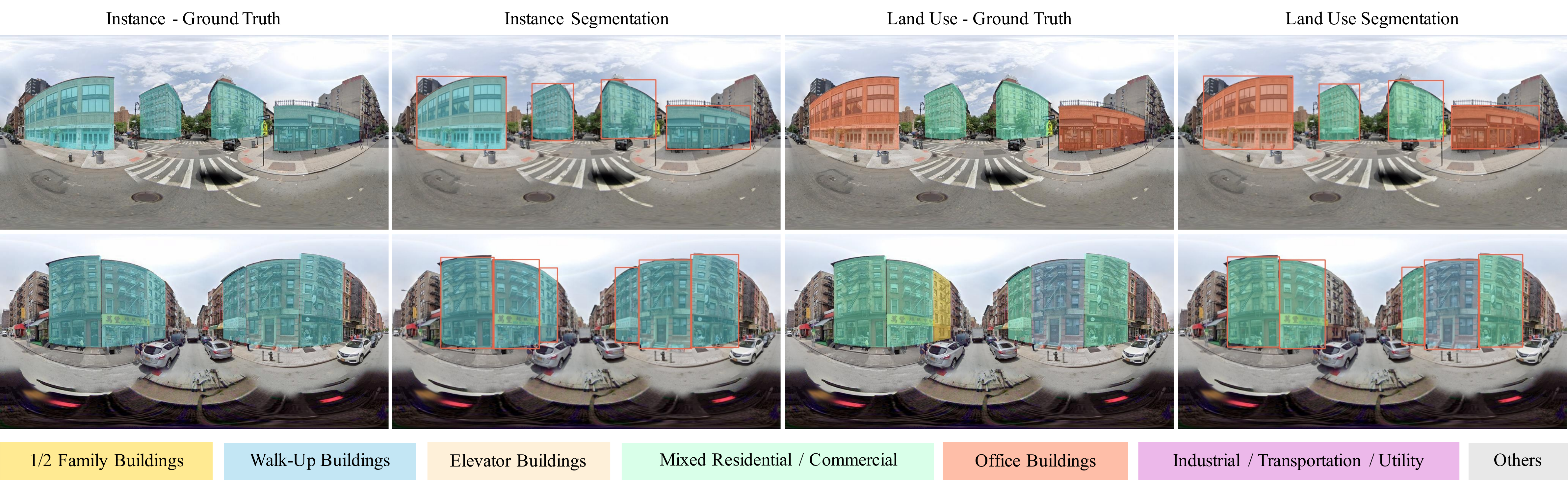}
\caption{Baseline results of street-level tasks (i.e., instance segmentation and land use segmentation) on panorama images. Different colors represent different land use categories.}
\label{fig:long}
\label{figure-panorama-2tasks}
\end{figure*}

\begin{figure}[t!]
\centering
\includegraphics[width=\linewidth]{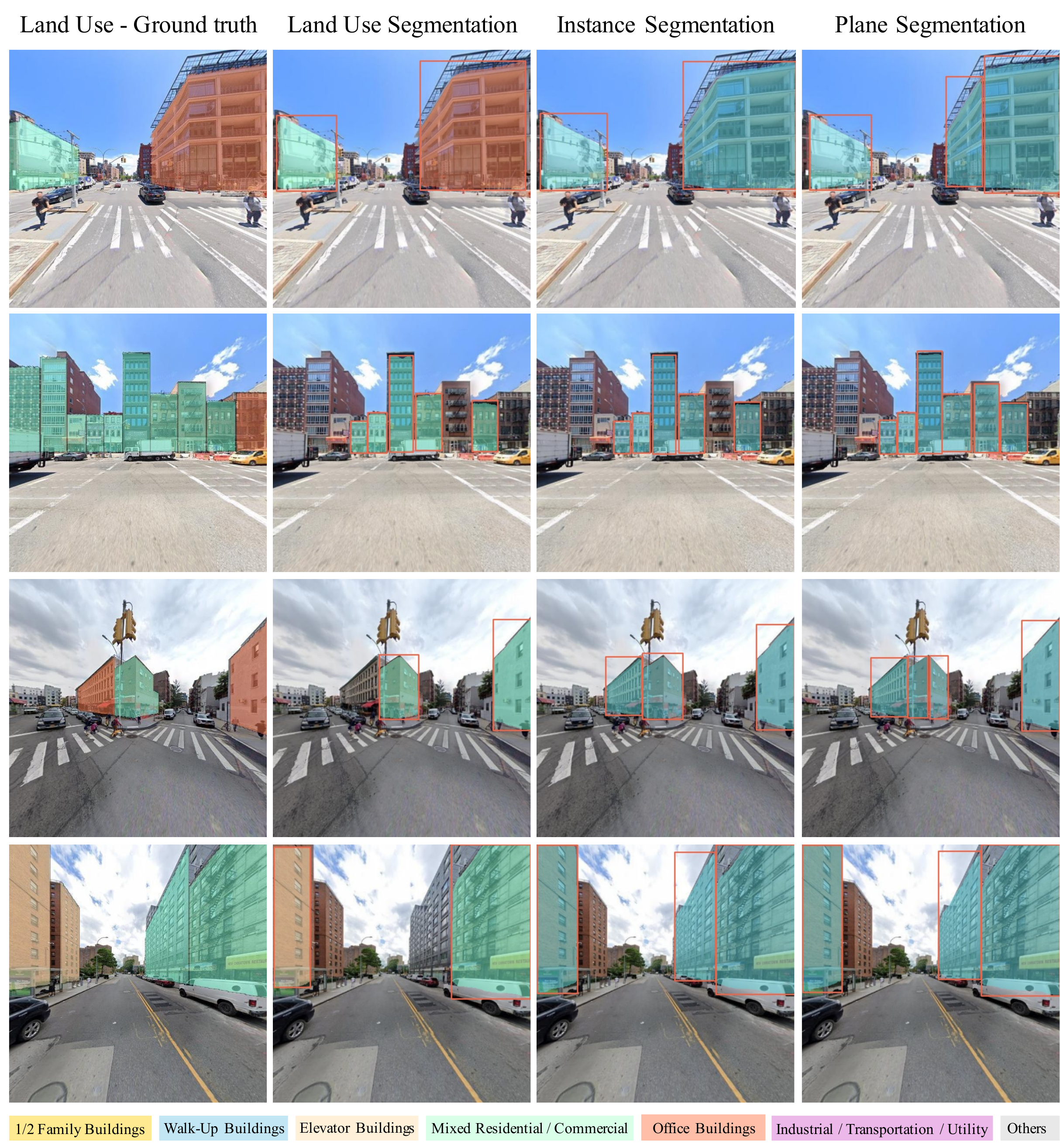}
\caption{Baseline results of street-level tasks (i.e., land use segmentation, instance segmentation and plane segmentation) on mono-view images.}
\label{fig:long}
\label{figure-monoview-3tasks}
\end{figure}

\subsection{Baselines for street-level tasks}\label{sec-baseline-street}

We analyze the performance of multiple segmentation tasks on street-level panorama and mono-view images. All street-level tasks are evaluated by COCO evaluation metrics, i.e., AP, $AP_{50}$, $AP_{75}$, $AP_{S}$, $AP_{M}$, and $AP_{L}$. As shown in Table \ref{table-panorama-2tasks} and Figure \ref{figure-panorama-2tasks}, the panorama-view images include a fine-grained land use segmentation task (denoted by Landuse Seg.) and a building instance segmentation task (denoted by Instance Seg.). Results show that the performance on instance segmentation task is significantly superior to the land use segmentation task under all metrics. 
Table \ref{table-monoview-3tasks} and Figure \ref{figure-monoview-3tasks} show the experimental results of mono-view images on three different tasks, i.e., landuse segmentation, instance segmentation, and plane segmentation. The buildings in the mono-view images contain either a single plane (e.g. the second and fouth rows of Figure \ref{figure-monoview-3tasks}) or multiple planes (the first and the third rows of Figure \ref{figure-monoview-3tasks}). Similar to the results of panorama-view images, the baseline method achieves much higher AP for the two binary segmentation tasks (plane and instance segmentation) compared with the fine-grained land use segmentation task.
From the qualitative results shown in Figure \ref{figure-panorama-2tasks} and Figure \ref{figure-monoview-3tasks}, we can find that the baseline method is capable of precisely extracting the instance and plane boundaries in most cases, but has difficulties in identifying the accurate land use type of some building instances. A detailed analysis of the performance on different land use types will be introduced in Section \ref{sec-different-methods}.

\linespread{1.2}
\begin{table}[!t]
\centering
\caption{Baseline results of street-level images from panorama view, in terms of COCO evaluation metrics.}
\label{table-panorama-2tasks}
\resizebox{\linewidth}{!}{%
\begin{tabular}{|c|c|c|c|c|c|c|}
\hline
Task           & $AP$ & $AP_{50}$ & $AP_{75}$ & $AP_S$ & $AP_M$ & $AP_L$ \\ \hline
Landuse Seg. & 26.0 & 34.7      & 28.5      & 0.3    & 12.0   & 30.4   \\ \hline
Instance Seg.  & 66.7 & 86.5      & 72.5      & 1.7    & 40.2   & 74.1   \\ \hline
\end{tabular}%
}
\end{table}

\begin{table}[!t]
\centering
\caption{Baseline results of street-level images from mono-view, in terms of COCO evaluation metrics.}
\label{table-monoview-3tasks}
\resizebox{\linewidth}{!}{%
\begin{tabular}{|c|c|c|c|c|c|c|}
\hline
Task           & $AP$ & $AP_{50}$ & $AP_{75}$ & $AP_S$ & $AP_M$ & $AP_L$ \\ \hline
Landuse Seg. & 23.9 & 32.1      & 26.7      & 0.3    & 10.6   & 27.5   \\ \hline
Instance Seg.  & 68.3 & 88.8      & 73.8      & 3.2    & 33.3   & 76.1   \\ \hline
Plane Seg.     & 65.1 & 87.4      & 71.0      & 5.0    & 40.7   & 73.8   \\ \hline
\end{tabular}%
}
\end{table}

\begin{table}[!t]
\centering
\caption{Land use and instance segmentation results of street-level images from panorama and mono-view, in terms of COCO evaluation metrics.}
\label{table-impact-view}
\resizebox{\linewidth}{!}{%
\begin{tabular}{|c|c|c|c|c|c|c|}
\hline
Task           & $AP$ & $AP_{50}$ & $AP_{75}$ & $AP_S$ & $AP_M$ & $AP_L$ \\ \hline
landuse-panorama & 20.6 & 31.0 & 23.0 & 0.1 & \textbf{11.6} & 24.8 \\ \hline
landuse-monoview & \textbf{23.9} & \textbf{32.1} & \textbf{26.7} & \textbf{0.3} & 10.6   & \textbf{27.5}   \\ \hline
instance-panorama & 61.6 & 84.0 & 67.5 & 0.5 & \textbf{38.9} & 70.4 \\ \hline
instance-monoview  & \textbf{68.3} & \textbf{88.8} & \textbf{73.8} & \textbf{3.2} & 33.3   & \textbf{76.1}   \\ \hline
\end{tabular}%
}
\end{table}

\begin{figure*}[t]
\centering
\includegraphics[width=\textwidth]{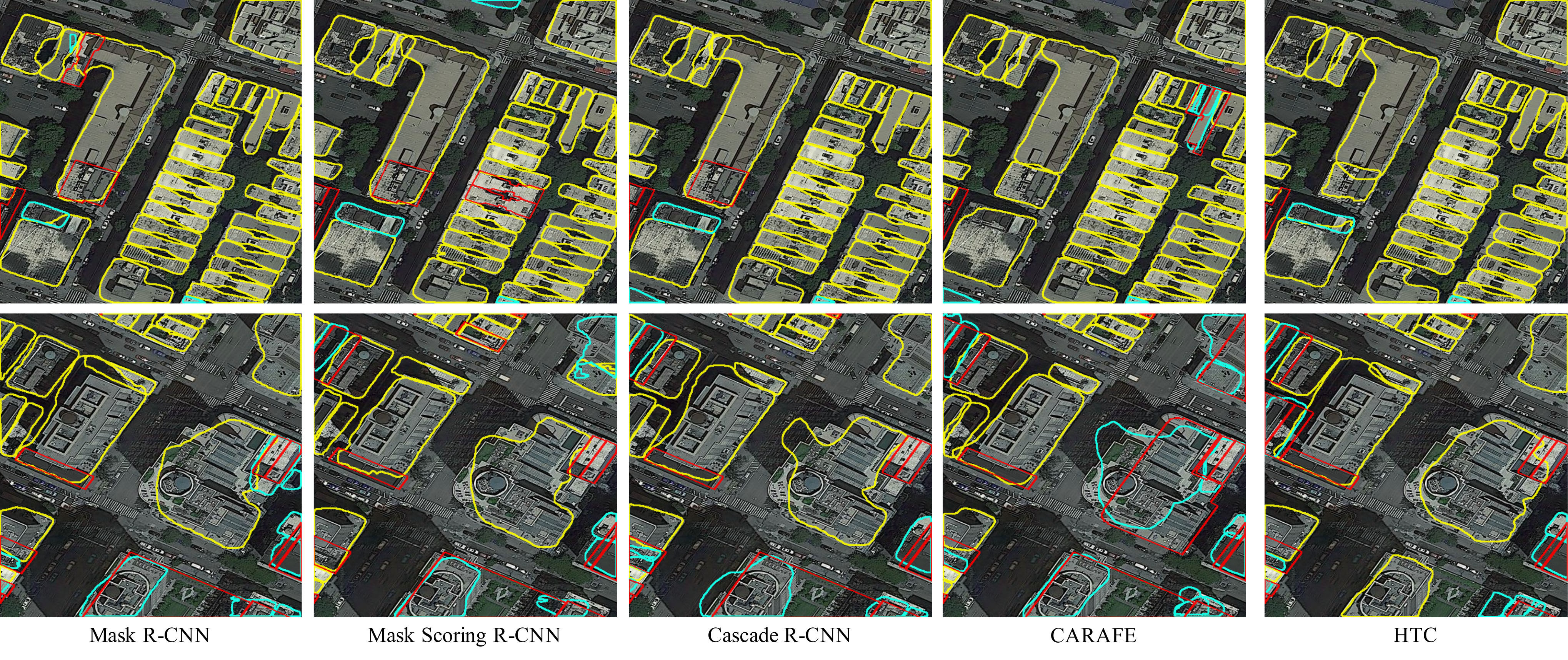}
\caption{Qualitative results of different methods for instance segmentation on satellite-level images. The yellow, cyan, and red polygons denote the TP, FP, and FN, respectively.}
\label{fig:long}
\label{figure-satseg-5methods}
\end{figure*}

\begin{table*}[h]
\centering
\caption{Quantitative results of different methods for instance segmentation on satellite-level images.}
\label{table-satseg-5methods}
\resizebox{.75\textwidth}{!}{%
\begin{tabular}{|c|c|c|c|c|c|c|c|c|c|}
\hline
\multirow{2}{*}{Method} & \multicolumn{6}{c|}{Metrics of various thresholds}                                            & \multicolumn{3}{c|}{threshold = 0.5} \\ \cline{2-10} 
                        & $AP$            & $AP_{50}$        & $AP_{75}$        & $AP_S$         & $AP_M$         & $AP_L$         & P      & R        & F1      \\ \hline
Mask R-CNN \cite{he2017mask}             & 29.7          & 66.0            & 23.5          & 15.9          & 33.9          & 36.7          & 76.9           & 66.3           & 71.2          \\ \hline
MS R-CNN  \cite{huang2019mask}     & 29.7          & 64.9          & 23.8          & 16.1          & 33.5          & 36.1          & \textbf{81.3}  & 62.3           & 70.5          \\ \hline
Cascade \cite{Cai_2019}     & 31.1          & 66.5          & 26.3          & 16.3          & 35.5          & 38.2          & 79.2           & 66.4           & 72.2          \\ \hline
CARAFE  \cite{wang2019carafe}                & 31.1          & 67.4          & 25.7          & \textbf{16.5} & 35.7          & 36.2          & 76.9           & \textbf{67.7}  & 72.0          \\ \hline
HTC  \cite{chen2019hybrid}                   & \textbf{32.3} & \textbf{68.7} & \textbf{27.2} & 16.3          & \textbf{36.6} & \textbf{38.8} & 78.9           & 67.2           & \textbf{72.6} \\ \hline
\end{tabular}%
}
\end{table*}

\section{Results analysis and discussions}
\label{sec-results}

\subsection{The impact of view on street-level tasks}

In Section \ref{sec-baseline-street}, we have analyzed the impact of view on satellite-level instance segmentation and height estimation tasks. In Table \ref{table-impact-view}, we further provide the land use and instance segmentation results of street-level images from panorama-view and mono-view. The experimental setting and performance of mono-view images are the same as those in Table \ref{table-panorama-2tasks} and Table \ref{table-monoview-3tasks}. To guarantee a fair comparison, we remove the images from the panorama-view dataset of which the geographical coordinates are not included in the mono-view dataset so that the two types of images are based on the same experimental setting and collection sites. Results show that the performance of mono-view images is superior to panorama images regarding $AP$, $AP_{50}$, $AP_{75}$, $AP_{S}$, and $AP_{L}$, with a slight drop on $AP_{M}$. The performance gap for the instance segmentation task is more obvious compared with the land use segmentation task, with an AP gap of 6.7\% and 3.3\%, respectively. The performance discrepancy between panorama-view and mono-view images might because the current baseline methods are widely evaluated on the common datasets (such as COCO and CityScapes) that are collected at a single view, without leveraging the special geometry properties of panorama images in network design. More analysis on the limitations of the existing methods will be provided in Section \ref{sec-limitations}.

\begin{figure*}[t!]
\centering
\includegraphics[width=\textwidth]{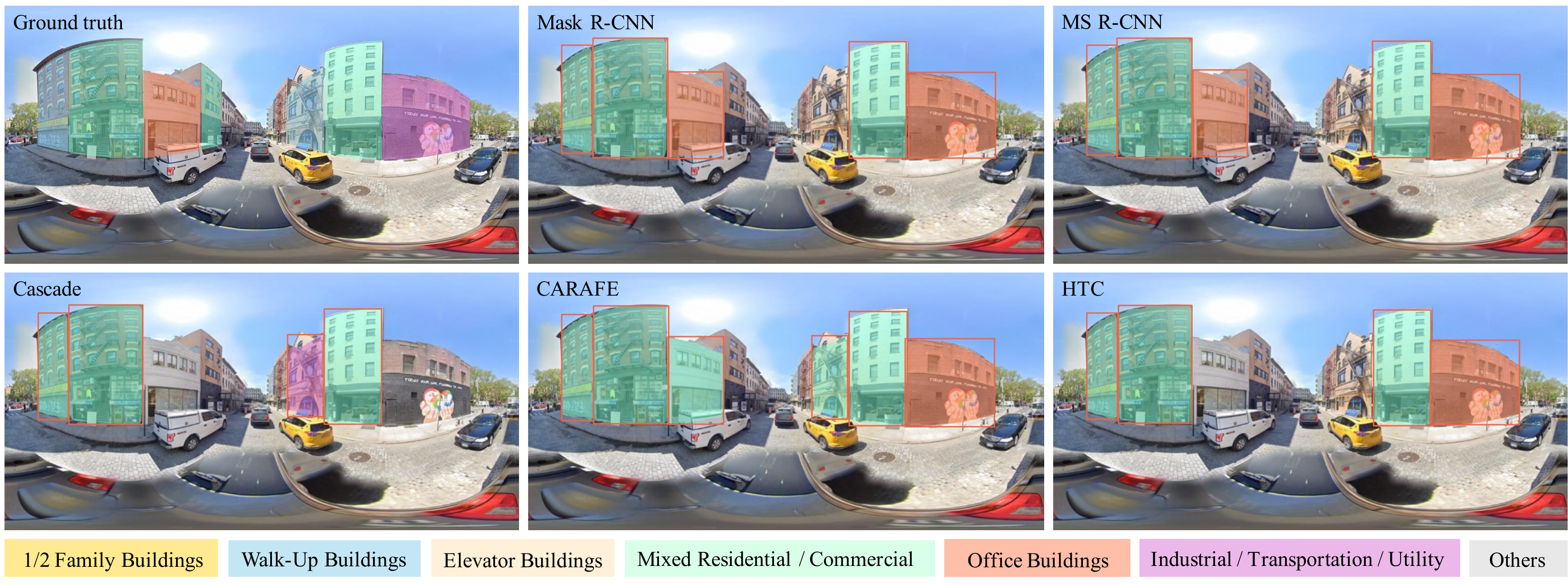}
\caption{Qualitative Results of different methods for fine-grained land use segmentation on street-level panorama images. Different colors represent different land use categories.}
\label{fig:long}
\label{figure-panorama-5methods}
\end{figure*}

\begin{table*}[!t]
\centering
\caption{Quantitative results of different methods for fine-grained land use segmentation on street-level panorama images.}
\label{table-panoseg-5methods}
\resizebox{\textwidth}{!}{%
\begin{tabular}{|c|c|c|c|c|c|c|c|c|c|c|c|c|c|}
\hline
\multirow{2}{*}{Method} & \multicolumn{6}{c|}{Overall Metrics}                                                         & \multicolumn{7}{c|}{Metrics of each category}                                                                \\ \cline{2-14} 
                        & $AP$          & $AP_{50}$     & $AP_{75}$     & $AP_S$       & $AP_M$        & $AP_L$        & C1            & C2            & C3            & C4            & C5            & C6            & C7           \\ \hline
Mask R-CNN \cite{he2017mask}             & 26.0          & 34.7          & 28.5          & \textbf{0.3} & 12.0          & 30.4          & 19.6          & 37.5          & 25.8          & 39.2          & 36.9          & 22.2          & 0.8          \\ \hline
MS R-CNN  \cite{huang2019mask}              & 27.1          & \textbf{35.8} & 29.8          & 0.1          & \textbf{12.4} & 31.5          & \textbf{22.5} & \textbf{39.1} & 26.2          & \textbf{40.8} & 38.0          & 21.7          & \textbf{1.2} \\ \hline
Cascade  \cite{Cai_2019}               & 25.9          & 33.8          & 28.3          & 0.2          & 11.4          & 30.5          & 20            & 38.3          & 25            & 38.5          & 36.7          & 22.1          & 0.3          \\ \hline
CARAFE \cite{wang2019carafe}                 & 25.9          & 34.5          & 28.5          & 0.1          & 11.9          & 30.2          & 19.6          & 37.3          & 24.9          & 39.9          & 37.2          & 21.5          & 0.8          \\ \hline
HTC  \cite{chen2019hybrid}                   & \textbf{27.2} & 35.7          & \textbf{29.9} & \textbf{0.3} & \textbf{12.4} & \textbf{32.0} & 20.8          & 38.7          & \textbf{27.2} & 39.9          & \textbf{38.4} & \textbf{24.5} & \textbf{1.2} \\ \hline
\end{tabular}%
}
\end{table*}


\subsection{Experimental results of different models}\label{sec-different-methods}

Besides the baseline results obtained from Mask R-CNN \cite{he2017mask}, we further provide the experimental results of other four commonly-used instance segmentation methods, including Mask Scoring R-CNN (denoted by MS R-CNN) \cite{huang2019mask}, Cascade Mask R-CNN (denoted by Cascade) \cite{Cai_2019}, Content-Aware ReAssembly of FEatures (denoted by CARAFE) \cite{wang2019carafe}, and Hybrid Task Cascade (denoted by HTC) \cite{chen2019hybrid}. All methods are implemented base on mmdetection \cite{chen2019mmdetection} with the same experimental setting described in Section \ref{sec-setting}. Table \ref{table-satseg-5methods} and Figure \ref{figure-satseg-5methods} show the building footprint segmentation results obtained from the above five methods on the satellite-level image dataset with a small view angle. Results demonstrate that the performance discrepancy between different methods is not as obvious as the one resulting from different view angles. Among the five methods, HTC achieves the best overall performance followed by Cascade and CARAFE for both COCO and SpaceNet evaluation metrics, indicating the effectiveness of the cascade structure and the feature reassembly scheme for building footprint segmentation.

Table \ref{table-panoseg-5methods} and Figure \ref{figure-panorama-5methods} show the land use segmentation results obtained from the five methods on street-level panorama images. Similar to the satellite-level building footprint segmentation task, HTC outperforms the other four methods in terms of both overall metrics and those of each category. Besides HTC, the MS R-CNN method also demonstrates promising performance for land use segmentation, which indicates that the mask scoring strategy can effectively improves the fine-grained segmentation performance of building instances. From the metrics of each category, we can find that C2, C4 and C5 (i.e., Walk-up Buildings, Mixed Residential/Commercial, and Office Buildings) have better performance compared with C1, C3 and C6 (i.e., 1/2 Family Buildings, Elevator Buildings and Industrial/Transportation/Utility) for all methods, with the AP ranging from 20\% to 40\%. The 'others' category has an extremely low AP score, which might resulting from the very small quantity of samples in this category. Compared with the residential buildings, the mixed Residential/Commercial and Office buildings have more special characteristics (i.e. the boundary between first and above floors, facade design, and building structure), contributing to the superior AP scores of these two categories. The performance of the above six categories is also concordant with the ratio of the sample quantity.

\subsection{Limitations of the current baselines on OmniCity}\label{sec-limitations}

In this section we summarize the limitations of existing methods for satellite and street-level tasks on our OmniCity dataset. For the building footprint segmentation task on the satellite imagery, the performance of existing methods get worse with the increasing of off-nadir view angle, which might due to the serious parallax and shadow effect on the satellite images with a large off-nadir view angle. For the height estimation task, most existing methods directly apply the monocular depth estimation methods to remote sensing scene and suffer from the following two shortcomings. These methods usually produce poor results on the invisible side of the footprint boundary, which can become much worse for high-rise buildings on very off-nadir images. On the other hand, existing methods use deep neural networks to regress continuous values following the depth estimation methods, while the ground truth and the desirable prediction of building heights are discrete values, resulting in difficulties for network training and extra efforts for converting the continuous prediction values into discrete height values via post-processing.
Several recent studies \cite{2020Learning,li20213d} propose novel methods to improve the height estimation and footprint segmentation performance for images with large off-nadir view angles. However, these methods require additional annotation efforts for labeling the offset for each building instance, i.e., the deviation between the roof and footprint. The required offset annotations are currently not available in our OmniCity dataset of which the satellite-level annotations are acquired from existing label maps and the manual annotations are only conducted on the street-level images. The offset annotations as well as the experimental results of these recent methods will be updated on OmniCity homepage in our future work.

For the street-level tasks based on panorama images, existing methods target at general instance segmentation tasks for commonly-used datasets, e.g. COCO \cite{lin2014microsoft}, CityScapes \cite{cordts2016cityscapes}, BDD100K \cite{yu2020bdd100k}, etc. For these datasets, the images often have a single capture view and a narrow Field of View (FoV). However, for panorama images, the special properties such as the wide FoV covering full 360-degree in the horizontal direction, are not taken into consideration in the design of existing methods, which results in the worse land use and instance segmentation performance of panorama images compared with the mono-view images to some extent.
In addition, for both mono-view and panorama images, existing methods have difficulties in accurately recognizing the building instance with a small area (e.g. buildings located in the side of the main parts), the land use categories with a small number of building instances (e.g. the category of the others), and the land use categories that are easily confused (e.g. 1/2 Family Buildings and Walk-Up Buildings, Mixed Residential/Commercial and Office Buildings). 
New instance segmentation methods should be designed for solving the above limitations considering the characteristics of panorama images, building instances, fine-grained land use categories, etc.

\section{Potential of the OmniCity dataset}
Our proposed OmniCity dataset demonstrates great potential for facilitating city understanding, machine perception, and generative modeling researches in many aspects. 

First, it can serve as a new dataset for the existing tasks such as ground-to-aerial image geo-localization, aerial-to-ground image synthesis, and segmentation/detection of buildings/trees/land use from cross-view images. For the existing datasets of the above tasks, the street-level images are not annotated for almost all datasets. Our OmniCity dataset, on the contrary, provides additional annotation types that are not contained in existing datasets, e.g. the building instances and land use categories on street-level images, which might promote novel methods to explore and leverage the new annotation types to improve the performance of these tasks. 
The annotations are organized in a unified version, which means multiple tasks can be performed on a single image, and thus can well support the multi-task learning setting. Additionally, since the building instances in our datasets are directly linked with the urban planning data using block-lot id, it is easy to further enrich our datasets with more types of annotations from other urban datasets, especially those for social and urban studies.

Second, our OmniCity dataset provides a new application scenario or problem setting to existing tasks. For line segment detection and wireframe parsing tasks, existing datasets contains densely distributed line segment and wireframe labels of indoor and outdoor scenes. For our OmniCity dataset, instead of labeling all line segments and wireframes on the building facades (e.g. windows and doors), we only annotate the main line segments and wireframes on the outlines of each building plane, which are much sparser compared with the existing datasets. The serious shelters from the trees and vehicles also bring challenges to these tasks. New line segment detection and wireframe parsing methods should be designed for the OmniCity scenario. 

Moreover, our OmniCity dataset facilitates new tasks for city reconstruction and simulation. 
Treating each panorama as a unique city scene, a complete 3D model representing such a local scene is available in our dataset, as also shown in the second window panel in our GUI (Figure~\ref{figure-annotation-tool}).
In addition, these 3D models are stored in abstract vector formats, with clean and clear vertices indicating the building facades, streets, etc, which are suitable for many shape generation tasks represented with graphs.

Finally, with the well-aligned satellite and street-level images as well as the various annotation types, a novel city reconstruction task, i.e., 3D building reconstruction from cross-view images, can be derived for producing holistic 3D buildings with both fine-grained land use category and precise geometry information (vector 3D model). 
Existing datasets and methods only target at height estimation or 3D reconstruction from monocular or multi-view remote sensing imagery.
For our OmniCity scenario, new methods should be designed to leverage the additional information from street-level images and annotations for improving the 3D reconstruction and semantic prediction performance.

\section{Conclusion}

In this paper, we have proposed OmniCity, a new dataset for omnipotent city understanding from satellite and street-level images of multiple views. The dataset contains over 100K images collected from 25K geo-locations in New York City, of which the annotations are generated from both existing label maps of satellite view and our proposed pipeline for efficient annotation of street-level images.
We provide baseline experimental results for multiple tasks based on state-of-the-art models, including building footprint extraction / height estimation on multi-view satellite images, and building plane / instance / land use segmentation on street-level panorama and mono-view images.
We also conduct comprehensive analysis regarding the impact of view on street-level tasks, the performance of different instance segmentation methods, limitations of the existing methods, etc.
We believe that OmniCity will not only promote new algorithms and provide new application scenarios for existing tasks, but facilitate novel tasks for 3D city reconstruction and simulation. 

In our future work, we will keep updating the OmniCity dataset and the benchmarks in the following aspects. Owing to the proposed annotation pipeline and the unified annotations with a vector format and rich meta information (geo-locations, block-lot id, etc.), OmniCity can be efficiently supplemented with more properties of buildings and other static geographical objects (roads, sidewalk, trees, open space, green space, etc.), and extended to other cities of different countries.
The benchmark results of more state-of-the-art models and new tasks will also be provided accordingly.
Finally, based on the rich annotation and view types of OmniCity, We plan to develop new methods for existing and novel tasks, such as object detection, instance segmentation, and 3D reconstruction from cross-view images.


%





\ifCLASSOPTIONcaptionsoff
  \newpage
\fi



\bibliographystyle{IEEEtran}
\bibliography{egbib}
%

%


\vskip -2.3\baselineskip plus -1fil

\begin{IEEEbiography}[{\includegraphics[width=1in,height=1.25in,clip,keepaspectratio]{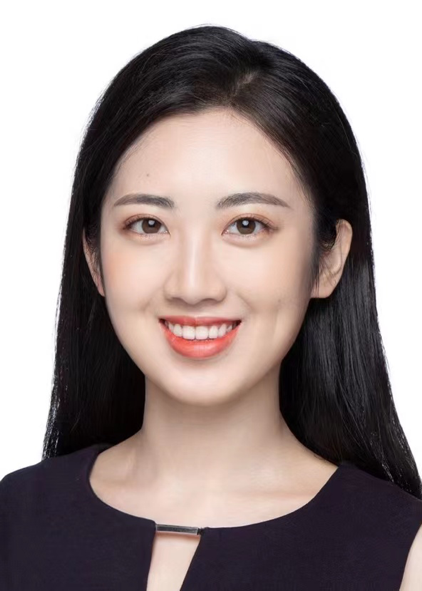}}]{Weijia Li}
is an Associate Professor at School of Geospatial Engineering and Science, Sun Yat-Sen University. From 2019 to 2021, she was a Post-doc Researcher at CUHK-Sensetime Joint Lab (MMLab), Department of Information Engineering, CUHK. She received the Ph.D. degree from Department of Earth System Science, Tsinghua Univeristy, in 2019, and the Bachelor degree from Department of Computer Science, Sun Yat-Sen Univeristy, in 2014. Her research interests include remote sensing image understanding, computer vision, and deep learning.
\end{IEEEbiography}

\vskip -2.3\baselineskip plus -1fil

\begin{IEEEbiography}[{\includegraphics[width=1in,height=1.25in,clip,keepaspectratio]{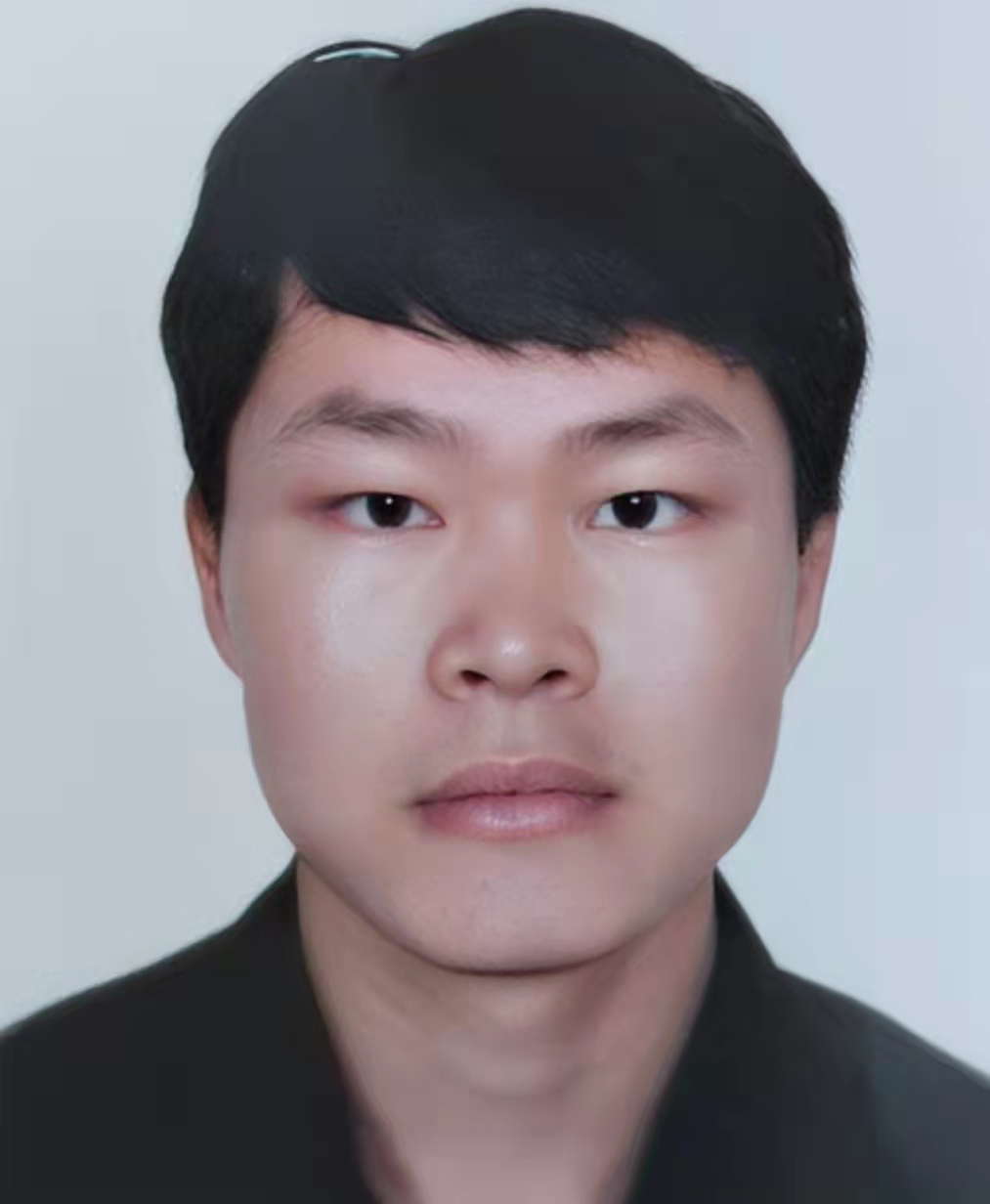}}]{Yawen Lai}
received the B.S. degree in computer science from the University of Electronic Science and Technology of China, in 2019. He is currently pursuing the M.S. degree in computer science with Peking University. His research interests include computer vision, multi-view stereo, depth estimation, and 3D reconstruction.
\end{IEEEbiography}

\vskip -2.3\baselineskip plus -1fil

\begin{IEEEbiography}[{\includegraphics[width=1in,height=1.25in,clip,keepaspectratio]{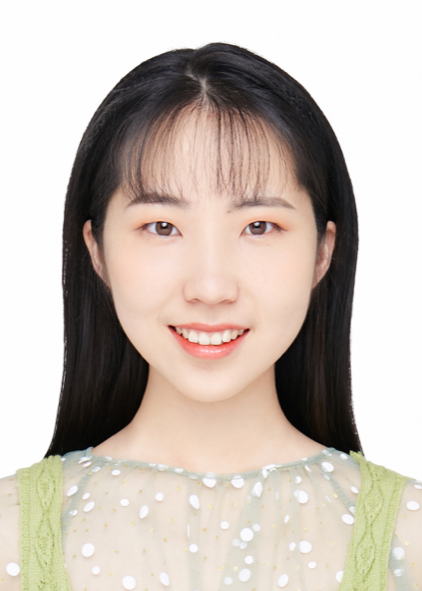}}]{Linning Xu}
is a Ph.D. student at Multimedia Lab, The Chinese University of Hong Kong.  Before that, she received the B.S. degree in Statistics from the Chinese University of Hong Kong, Shenzhen in 2020. She has broad interest in Machine Learning and Computer Vision. Her current research focuses on multiple interdisciplinary areas including AI + city and AI + movie.
\end{IEEEbiography}

\vskip -2.3\baselineskip plus -1fil

\begin{IEEEbiography}[{\includegraphics[width=1in,height=1.25in,clip,keepaspectratio]{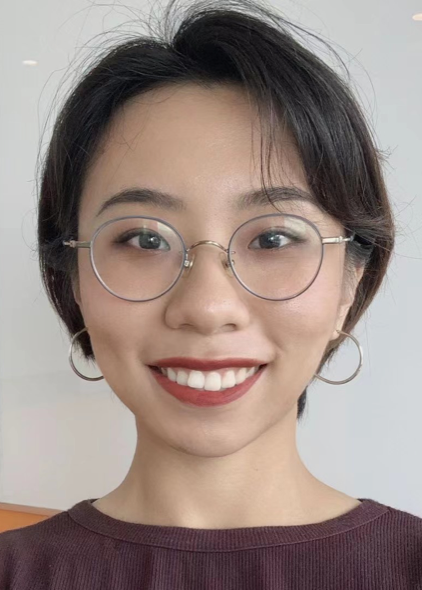}}]{Yuanbo Xiangli} 
is currently a third year PhD student at MMLab, Department of Information Engineering, CUHK. She received the Bachelor degree from the University of Nottingham and a Master degree from Oxford University. Her research interests are majorly in generative modeling, multi-modality learning and 3D vision.
\end{IEEEbiography}

\vskip -2.3\baselineskip plus -1fil

\begin{IEEEbiography}[{\includegraphics[width=1in,height=1.25in,clip,keepaspectratio]{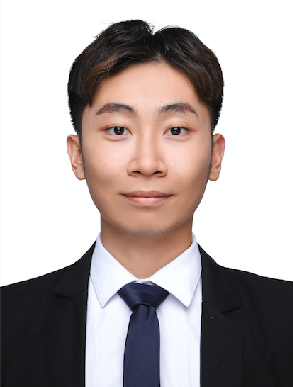}}]{Jinhua Yu}
received the B.S. degree in geomatics engineering from Nanjing University of Posts and Telecommunications, in 2022. He is currently pursuing the M.S. degree at School of Geospatial Engineering and Science, Sun Yat-Sen University. His research interests include remote sensing image understanding, computer vision, and deep learning.
\end{IEEEbiography}

\vskip 3.3\baselineskip plus -1fil

\begin{IEEEbiography}[{\includegraphics[width=1in,height=1.25in,clip,keepaspectratio]{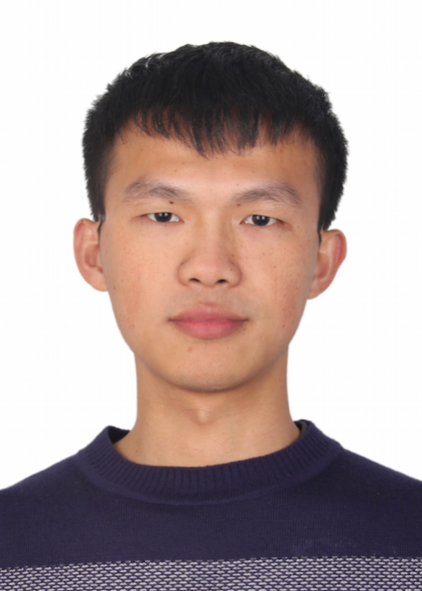}}]{Conghui He}
is a research director in SenseTime. He received the Bachelor and Ph.D. degree in Sun Yat-sen University (2013) and Tsinghua University (2018), respectively. His research interests include high performance computing, reconfigurable computing, graph computing, and computer vision. He won the 2017 ACM Gordon Bell Prize award and the outstanding Ph.D. award in Tsinghua University.
\end{IEEEbiography}

\vskip -2.3\baselineskip plus -1fil

\begin{IEEEbiography}[{\includegraphics[width=1in,height=1.25in,clip,keepaspectratio]{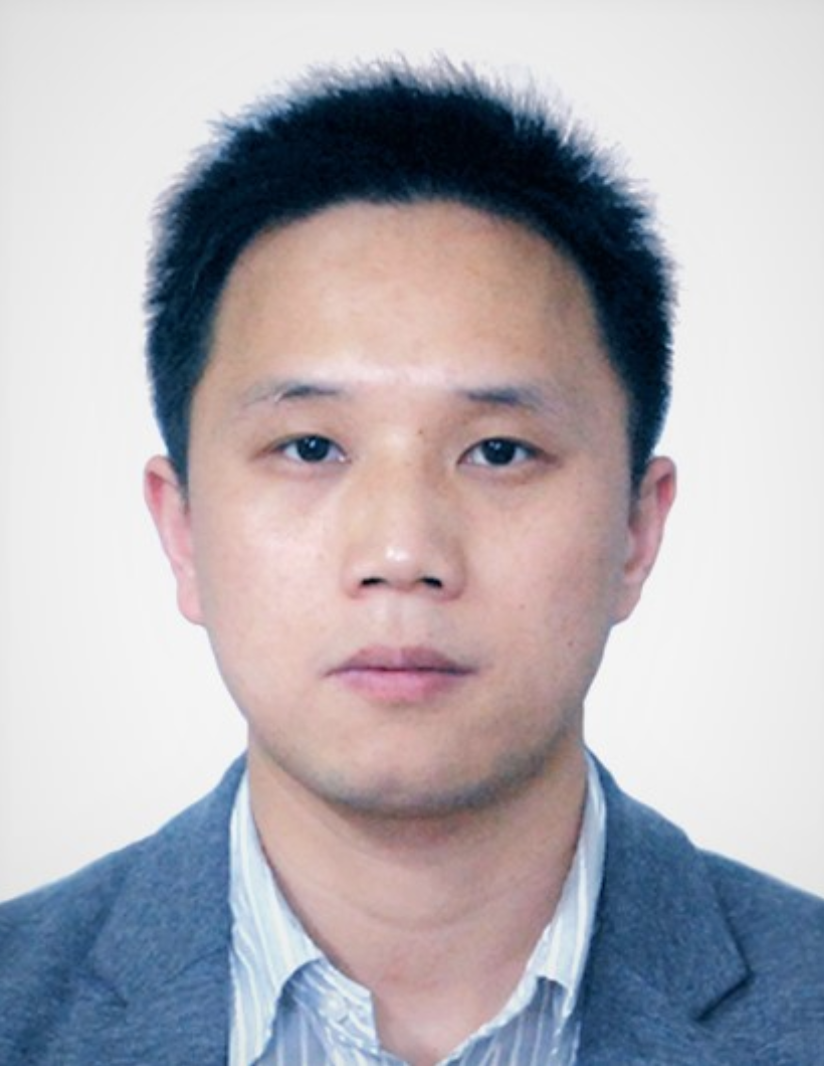}}]{Gui-Song Xia}
received his Ph.D. degree in image processing and computer vision from CNRS LTCI, T{\'e}l{\'e}com ParisTech, Paris, France, in 2011. From 2011 to 2012, he has been a Post-Doctoral Researcher with the Centre de Recherche en Math{\'e}matiques de la Decision, CNRS, Paris-Dauphine University, Paris, for one and a half years.
He is currently working as a full professor in computer vision and photogrammetry at Wuhan University. He has also been working as Visiting Scholar at DMA, {\'E}cole Normale Sup{\'e}rieure (ENS-Paris) for two months in 2018. He is also a guest professor of the Future Lab AI4EO in Technical University of Munich (TUM). His current research interests include mathematical modeling of images and videos, structure from motion, perceptual grouping, and remote sensing image understanding. He serves on the Editorial Boards of several journals, including {\em ISPRS Journal of Photogrammetry and Remote Sensing, Pattern Recognition, Signal Processing: Image Communications, EURASIP Journal on Image \& Video Processing, Journal of Remote Sensing, and Frontiers in Computer Science: Computer Vision}.
\end{IEEEbiography}

\vskip -2.3\baselineskip plus -1fil

\begin{IEEEbiography}[{\includegraphics[width=1in,height=1.25in,clip,keepaspectratio]{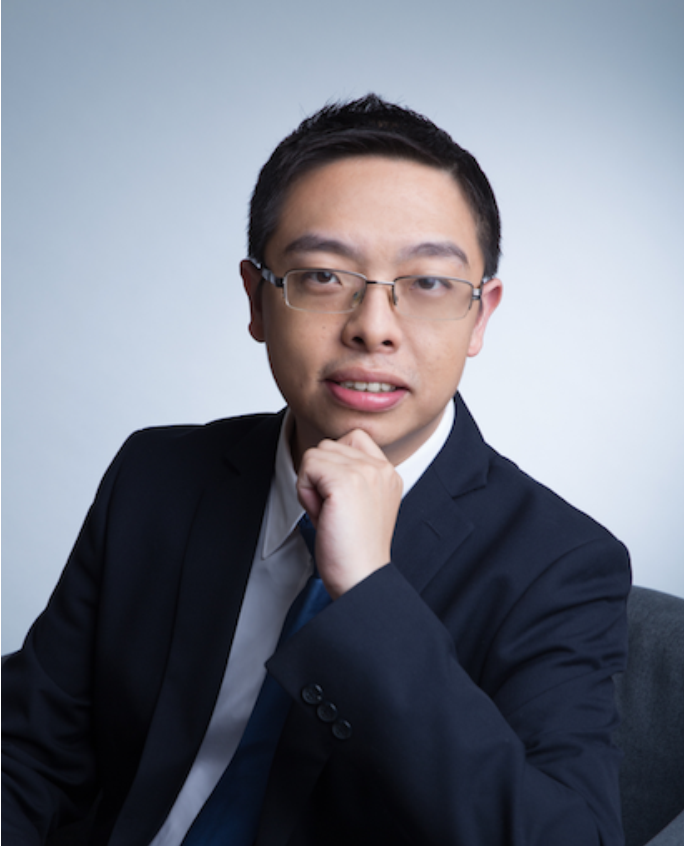}}]{Dahua Lin} 
is an Associate Professor at the department of Information Engineering, the Chinese University of Hong Kong, and the Director of CUHK-SenseTime Joint Laboratory. He received the B.Eng. degree from the University of Science and Technology of China (USTC) in 2004, the M. Phil. degree from the Chinese University of Hong Kong (CUHK) in 2006, and the Ph.D. degree from Massachusetts Institute of Technology (MIT) in 2012. Prior to joining CUHK, he served as a Research Assistant Professor at Toyota Technological Institute at Chicago, from 2012 to 2014. His research interest covers computer vision and machine learning. He serves on the editorial board of the International Journel of Computer Vision (IJCV). He also serves as an area chair for multiple conferences, including ECCV 2018, ACM Multimedia 2018, BMVC 2018, CVPR 2019, BMVC 2019, AAAI 2020, and CVPR 2021.
\end{IEEEbiography}






\end{document}